\documentclass[times, review, 10pt]{elsarticle}
\usepackage{amsmath,amssymb,amsfonts,amsthm,mathtools,bm}
\usepackage{booktabs,multirow,array,tabularx}
\usepackage[ruled,vlined,linesnumbered]{algorithm2e}
\usepackage{graphicx}
\usepackage[hidelinks]{hyperref}
\usepackage{enumitem}
\usepackage{microtype}
\usepackage{setspace}
\usepackage{xcolor}
\usepackage{newtxtext}      
\usepackage{newtxmath}
\newtheorem{theorem}{Theorem}
\newtheorem{proposition}{Proposition}

\newcommand{\R}{\mathbb{R}}
\newcommand{\one}{\bm 1}

\newcommand{\by}{\bm y}
\newcommand{\bw}{\bm w}
\newcommand{\br}{\bm r}
\newcommand{\bu}{\bm u}
\newcommand{\ba}{\bm\alpha}
\newcommand{\cT}{\mathcal T}
\newcommand{\diag}{\operatorname{diag}}
\newcommand{\proj}{\operatorname{proj}}
\newcommand{\argmin}{\operatorname*{arg\,min}}

\newcommand{\norm}[1]{\left\lVert#1\right\rVert}
\newcommand{\soft}{\mathcal S}

\newcolumntype{Y}{>{\raggedright\arraybackslash}X}

\journal{Pattern Recognition}

\begin{document}
\doublespacing
\begin{frontmatter}
\title{Data-Driven Pinball-Loss Selection for Vertically Distributed Elastic-Net SVMs}
\author[inst3]{Xiaofei Wu\fnref{equal}}
\author[inst2]{Kai Qi\fnref{equal}}
\author[inst1]{Rongmei Liang\corref{cor1}}
\fntext[equal]{These authors contributed equally to this work.}
\cortext[cor1]{Corresponding author: Rongmei Liang (liang\_r\_m@163.com).}
\address[inst3]{Yunnan Key Laboratory of Statistical Modeling and Data Analysis, School of Mathematics and Statistics, Yunnan University, East Outer Ring South Road, University Town, Chenggong District, Kunming, Yunnan, China}
\address[inst2]{National Center for Applied Mathematics in Chongqing, Chongqing Normal University, No. 37 University Town Middle Road, Shapingba District, Chongqing, China}
\address[inst1]{Department of Statistics and Data Science, Southern University of Science and Technology, 1088 Xueyuan Avenue, Nanshan District, Shenzhen, Guangdong 518055, China}

\begin{abstract}
The pinball-loss support vector machine is robust, but its asymmetry parameter is usually fixed in advance. We propose a data-driven elastic-net support vector machine that learns simplex-constrained weights over candidate pinball losses while retaining one classifier. The weighted loss is equivalent to a pinball loss with a data-dependent effective parameter. An empirical oracle inequality shows that, when weight regularization and simplex truncation vanish, the classifier objective at a global minimizer does not exceed that of the best fixed candidate; otherwise, the excess is explicitly bounded. For high-dimensional data, we develop a column-partitioned variable-splitting solver. It converges with a best-iterate $O(1/T)$ squared-step residual rate. Under common initialization and global parameters, any column partition produces, in exact arithmetic, the same iterates and solution as centralized training. Experiments assess predictive behavior, numerical equivalence, and multi-process scalability.
\end{abstract}

\end{frontmatter}

\section{Introduction}
Consider a binary training sample $\{(\bm x_i,y_i)\}_{i=1}^n$, where $n$ is the sample size, $\bm x_i\in\R^p$ is the $p$-dimensional feature vector for observation $i$, and $y_i\in\{-1,1\}$ is its class label. For a generic input $\bm x\in\R^p$, a regularized support vector machine estimates
\[
f(\bm x)=\langle \bw,\phi(\bm x)\rangle_{\mathcal H}+b,
\]
where $f$ is the decision function, $\phi:\R^p\to\mathcal H$ is a feature map, $\mathcal H$ is the induced feature space, $\langle\cdot,\cdot\rangle_{\mathcal H}$ is its inner product, $\bw\in\mathcal H$ is the coefficient element, and $b\in\R$ is the intercept. The identity map yields a linear model, whereas a kernel-induced map yields a nonlinear one. Because the present study targets high-dimensional, often sparse, data, we focus on linear SVMs, which offer scalable training, interpretable coefficients, and a natural basis for elastic-net regularization and vertical feature splitting \cite{joachims1998text,joachims2006linear,fan2008liblinear}. Hence, $\phi(\bm x)=\bm x$ throughout the remainder of the paper.

Within this linear framework, the fitted classifier is determined jointly by the margin loss and the regularizer. The hinge loss is computationally convenient but has a fixed influence pattern \cite{bartlett2006classification}. The pinball loss introduces an asymmetry parameter $\tau\in[0,1]$, which can improve robustness while preserving convexity \cite{huang2014pinsvm}, whereas elastic-net regularization promotes sparsity and stabilizes correlated features \cite{zou2005elastic,wang2006drsvm}. Combining these components is especially attractive in high-dimensional problems, where the number of features may be comparable to or much larger than the sample size. As the feature dimension grows, storing the complete design matrix and repeatedly evaluating global matrix--vector products on one machine can become impractical. When feature blocks are naturally held by different sites or exceed the memory of a single machine, column partitioning is therefore a practical requirement rather than merely an implementation choice. These considerations lead to two connected questions: how should $\tau$ be selected from the data, and how can the resulting model be trained efficiently and exactly when the features are partitioned by columns?

The first question concerns data-driven selection of $\tau$. A conventional grid search fits an independent model for each candidate value and ultimately returns only a discrete choice. We instead learn simplex weights over a finite family of normalized pinball losses while sharing one classifier and one residual vector $\br=(r_1,\ldots,r_n)^\top\in\R^n$, whose $i$th entry is the margin residual of observation $i$. The weighted family is exactly equivalent to a single normalized pinball loss with a data-dependent effective parameter. This representation also yields a direct oracle comparison. Specifically, let $\gamma\geq0$ denote the coefficient of the quadratic weight penalty and let $\varepsilon\geq0$ denote the lower bound imposed on each simplex weight. At any global minimizer, the empirical classifier objective is no larger than that of the best fixed candidate when $\gamma=\varepsilon=0$; otherwise, the excess is controlled by explicit weight-regularization and simplex-truncation terms.

The second question concerns computation for vertically partitioned high-dimensional data. Let $\bm X\in\R^{n\times p}$ be the design matrix, let $\by=(y_1,\ldots,y_n)^\top\in\R^n$ be the label vector, and let $\bw\in\R^p$ be the linear coefficient vector. The feature columns of $\bm X$ and the corresponding blocks of $\bw$ are distributed across $M$ workers, where $M$ is the number of feature-owning sites, while a coordinator maintains the shared intercept and sample-wise variables. By exchanging only partial margins, the proposed variable-splitting algorithm reproduces the centralized iterates and solution under any column partition in exact arithmetic, provided that the initialization and global algorithmic parameters are held fixed \cite{dai2022verticox,wu2025featureppa}.

To address these two questions jointly, the paper makes two contributions.
\begin{enumerate}[leftmargin=1.6em,label=(\arabic*)]
\item We propose a shared-classifier elastic-net SVM that learns simplex weights over normalized pinball-loss candidates. The weighted loss has an exact data-dependent effective parameter. An empirical oracle inequality shows that, when weight regularization and simplex truncation vanish, the classifier objective at a global minimizer does not exceed that of the best fixed-$\tau$ candidate; otherwise, the excess is explicitly bounded.
\item We develop a column-partitioned variable-splitting algorithm that distributes feature and coefficient blocks while reconstructing the global margin from local contributions. The inner solver converges with a best-iterate $O(1/T)$ rate for the squared step residual and is insensitive to the column partition: under common initialization and global parameters, every partition reproduces the centralized iterates and solution in exact arithmetic.
\end{enumerate}

The remainder of the paper is organized as follows. Section~2 reviews robust SVM losses, data-driven loss weighting, and vertical feature partitioning. Section~3 introduces the shared-classifier loss-weighting model and its effective-parameter representation. Section~4 develops the vertically distributed solver and analyzes its communication pattern. Section~5 presents the oracle comparison, descent, convergence, and partition-equivalence results. Section~6 reports the predictive, parameter-selection, numerical-invariance, and multi-process experiments. Section~7 concludes the paper, and the accompanying Supplementary Material provides detailed proofs and additional technical derivations. The replication code for the paper is available for download at \url{https://github.com/xfwu1016/DP-ENSVM}.

\section{Related work}
\subsection{Pinball loss and elastic-net SVMs}
The hinge loss is convex, classification calibrated, and compatible with efficient large-margin optimization, but its piecewise-linear geometry treats all observations at or inside the margin with the same slope. Robust alternatives modify this influence pattern while preserving convexity or computational tractability. Among them, the pinball loss assigns different slopes to the two sides of the margin residual and admits a quantile-oriented interpretation \cite{huang2014pinsvm}. The C-loss and rescaled hinge loss provide related robust alternatives that reshape the influence of difficult observations \cite{singh2014closs,xu2017rescaled}. The truncated pinball loss restores sparsity while reducing sensitivity to feature noise \cite{shen2017truncated}, and a bounded exponential-quantile construction further controls the influence of extreme observations \cite{li2024bounded}. Recent studies have also developed generalized-ramp, capped-squared, and wave losses together with scalable or smooth optimization procedures \cite{wang2024ramp,wang2024capped,akhtar2024wave}. Robust probability machines with embedded elastic-net feature selection provide a complementary view of the interaction between robustness and sparsity \cite{carrasco2025embedded}. These methods enrich the available loss catalogue, but they generally fix the selected loss and its shape parameters before model fitting. The present study addresses the distinct problem of data-driven selection within a finite normalized pinball-loss family.

The parameter $\tau$ determines the asymmetric geometry of the empirical loss and therefore plays a different role from an ordinary regularization parameter. While regularization controls model complexity, sparsity, and stability, $\tau$ changes the relative influence of residuals on the two sides of the decision margin. Hence, different values of $\tau$ may lead to distinct classifiers even under the same regularization setting. This motivates learning candidate weights explicitly and reporting the selected $\widehat\tau$ rather than treating $\tau$ as an invisible numerical setting.

Elastic-net regularization combines an $\ell_1$ absolute-value penalty, which promotes sparsity, with a strictly convex squared-$\ell_2$ Euclidean penalty, producing sparse and stable slope estimates in high-dimensional SVMs. The resulting convex composite problems can be solved by proximal-point, primal--dual, and augmented-Lagrangian techniques \cite{cai2013ppa,gu2014customized,he2020optimal,chen2016multiblock}. Liang et al.\ developed a linearized ADMM method for elastic-net SVMs, including pinball-loss models \cite{liang2024ladmm}. In the vertically distributed formulation, the slope vector is partitioned according to feature blocks, while the SVM intercept remains a single scalar maintained by the coordinator.  However, the linearized ADMM in \cite{liang2024ladmm} does not directly yield the feature-block-separable updates and partition-equivalence property required here.

\subsection{Data-driven loss weighting}
Learning convex combinations of candidate structures is common in kernel learning, graph-based learning, and classifier ensembles. Multiple-kernel learning estimates weights over representations \cite{gonen2013lmkl}, whereas prediction-level ensembles combine already fitted decision rules. Ensemble manifold regularization is especially relevant because it learns simplex weights over candidate graph Laplacians and alternates between the predictive model and the structural weights \cite{geng2012emr}. A quadratic penalty on the weight vector prevents an unstable winner-take-all solution and allows several candidates to remain active.

Our use of weights differs from a prediction ensemble. The candidates are loss geometries, not separate decision functions. One shared classifier produces one margin residual, and every candidate loss is evaluated on that same residual. The resulting convex combination has an exact effective-$\tau$ representation. Consequently, the weights summarize how the data distribute support over the candidate loss family; they do not average predictions and they do not create candidate-specific classifier parameters.

This distinction also separates the proposed method from ordinary grid search. Grid search fits independent models and makes a discrete decision using validation performance. Here the candidate risks enter one joint objective, the weights are updated from the current shared residual, and the effective parameter changes during optimization. The final discrete value $\widehat\tau$ is obtained only after the weighting process has stabilized.

\subsection{Vertical feature partitioning}
Vertical partitioning means that sites share aligned sample identifiers but own disjoint feature blocks. This differs from horizontal federated learning, in which sites possess different observations with a common feature schema. Recent federated-learning studies address dynamic client heterogeneity and personalized graph-structured information exchange \cite{guo2024fedmlp,rasti2025fedpnp}. Those settings are primarily organized around distributed observations or client models, whereas vertical learning must reconstruct one global prediction from disjoint feature views. Multi-party vertical frameworks with a coordinating server and vertical feature-selection methods both rely on exchanging intermediate representations rather than raw feature tables \cite{anees2024vfl,feng2022vflfs}. For a linear model, the global margin can be assembled especially transparently from local contributions. VERTICOX demonstrates this principle for survival analysis: each institution keeps its covariates locally, computes an intermediate score, and communicates aggregated quantities to a server \cite{dai2022verticox}.

The same architecture applies to the shared SVM classifier. With $M$ feature sites, site $m$ stores the local design block $\bm X_m$ and coefficient block $\bw_m$, where $m=1,\ldots,M$. Letting $\bm Y=\diag(\by)$, the site computes $\bm h_m=\bm Y\bm X_m\bw_m$ and returns $\bm h_m\in\R^n$ to the coordinator. The coordinator forms $\sum_m\bm h_m$, updates the common residual and dual vector, and maintains the scalar intercept $b$. There is no $b_m$ because the intercept is not associated with any feature subset.

Parallel optimization for high-dimensional regularized models can be organized through either consensus decompositions or direct feature splitting. Wu et al.\ developed a unified consensus-based parallel method for combined regularizations \cite{wu2025consensus}, and later introduced a feature-splitting proximal-point algorithm whose iterates are insensitive to the column partition \cite{wu2025featureppa}. The latter shared-vector structure is adapted here to the elastic-net PinSVM subproblem. Table~\ref{tab:related} positions the resulting method relative to neighboring research lines.

\begin{table*}[t]
\centering
\caption{Positioning relative to representative research lines.}
\label{tab:related}
\scriptsize
\setlength{\tabcolsep}{4pt}
\begin{tabularx}{\textwidth}{@{}YY>{\centering\arraybackslash}p{0.13\textwidth}>{\centering\arraybackslash}p{0.14\textwidth}@{}}
\toprule
Research line & Learned object & Data-driven $\tau$ & Column splitting \\
\midrule
Robust SVM losses \cite{huang2014pinsvm,singh2014closs,xu2017rescaled,shen2017truncated} & One fixed-loss classifier & No & No \\
Multiple-kernel learning \cite{gonen2013lmkl} & Kernel combination weights & No & Usually no \\
Structure-weighting methods \cite{geng2012emr} & Candidate-structure weights & Indirect & Usually no \\
Elastic-net SVM solvers \cite{wang2006drsvm,liang2024ladmm} & One fixed-loss sparse classifier & No & Not intrinsic \\
Proposed method & Loss weights and one shared classifier & Yes & Yes \\
\bottomrule
\end{tabularx}
\end{table*}

\section{Data-driven pinball-loss selection}
\subsection{Candidate loss family and shared residual}
\begin{figure}[t]
\centering
\includegraphics[width=.72\linewidth]{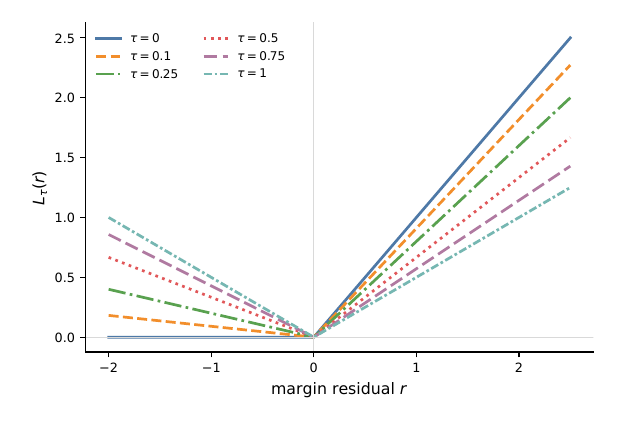}
\caption{Normalized pinball losses for $\cT=\{0,0.1,0.25,0.5,0.75,1\}$.}
\label{fig:loss}
\end{figure}

Let $\{(\bm x_i,y_i)\}_{i=1}^n$ be a binary training sample, where $n$ is the sample size, $\bm x_i\in\R^p$ is the feature vector of observation $i$, $p$ is the number of features, and $y_i\in\{-1,1\}$ is its class label. Let $\bm X\in\R^{n\times p}$ be the design matrix, $\by=(y_1,\ldots,y_n)^\top$ the label vector, $\bm Y=\diag(\by)$ the diagonal label matrix, and $\bar{\bm X}=\bm Y\bm X$ the label-weighted design matrix. For a scalar margin residual $r\in\R$ and asymmetry parameter $\tau\in[0,1]$, the normalized pinball loss is
\[
L_\tau(r)=
\begin{cases}
r/(1+\tau), & r\geq0,\\
-\tau r/(1+\tau), & r<0,
\end{cases}
\qquad \tau\in[0,1].
\]
The normalization makes the two absolute slopes sum to one, so losses with different $\tau$ values are comparable in scale. Let $\cT=\{\tau_1,\ldots,\tau_G\}$ denote the finite candidate set, where $G$ is its cardinality. Figure~\ref{fig:loss} shows the candidate family used in the experiments.

The classifier and residual are shared across all candidates:
\[
f(\bm x)=\bm x^\top\bw+b,
\qquad
\br=\one-\by b-\bar{\bm X}\bw.
\]
For each $\tau_g\in\cT$, define a loss function on this same residual vector,
\begin{equation}
\ell_g(\br)
=\frac1n\sum_{i=1}^nL_{\tau_g}(r_i),
\qquad g=1,\ldots,G.
\label{eq:candidaterisk}
\end{equation}
The index $g$ labels the loss geometry through $\tau_g$; it does not label a residual variable. Hence the model contains one $b$, one $\bw$, and one $\br$, and every candidate risk is evaluated at the same margin residual.

\subsection{Joint weighting model}
Let $\ba=(\alpha_1,\ldots,\alpha_G)^\top\in\R^G$ be the candidate-weight vector. For a prescribed lower-bound parameter $\varepsilon\geq0$, define the truncated simplex
\[
\Delta_G^\varepsilon=
\left\{\ba\in\R^G:\bm 1_G^\top\ba=1,
\ \alpha_g\geq\varepsilon\right\},
\qquad 0\leq G\varepsilon<1.
\]
Here $\bm 1_G\in\R^G$ is the all-ones vector. Let $\lambda_1\geq0$ be the sparsity-penalty parameter, $\lambda_2>0$ the quadratic regularization parameter, and $\gamma\geq0$ the weight-dispersion parameter. For a vector $\bm v$, $\norm{\bm v}_1$ denotes the sum of the absolute values of its entries and $\norm{\bm v}_2$ denotes its Euclidean norm. The proposed optimization model is
\begin{equation}
\begin{aligned}
\min_{b,\bw,\br,\ba}\quad
&\sum_{g=1}^G\alpha_g\ell_g(\br)
+\lambda_1\norm{\bw}_1
+\frac{\lambda_2}{2}\norm{\bw}_2^2
+\frac{\gamma}{2}\norm{\ba}_2^2,\\
\text{s.t.}\quad
&\by b+\bar{\bm X}\bw+\br=\one,\\
&\ba\in\Delta_G^\varepsilon.
\end{aligned}
\label{eq:model}
\end{equation}
The candidate index labels the family of loss functions $\ell_g$, each assigned a simplex weight $\alpha_g$; the residual $\br$ is a single shared optimization variable and carries no candidate index. Accordingly, the margin constraint and elastic-net penalty are each written once. The intercept is not regularized, following the usual linear SVM convention, while $\gamma\norm{\ba}_2^2/2$ controls the concentration of the candidate weights.

The shared-residual structure yields an exact representation.

\begin{proposition}[Effective pinball-loss representation]\label{prop:effective}
Let $\ba\in\Delta_G^\varepsilon$ and let $r$ be common to all candidate losses. Then
\begin{equation}
\sum_{g=1}^G\alpha_gL_{\tau_g}(r)
=L_{\tau_{\mathrm{eff}}}(r),
\qquad
\tau_{\mathrm{eff}}
=\frac{\sum_g\alpha_g\tau_g/(1+\tau_g)}
{\sum_g\alpha_g/(1+\tau_g)}.
\label{eq:taueff}
\end{equation}
Here $\tau_{\mathrm{eff}}$ denotes the effective pinball parameter induced by the candidate weights. Consequently, for fixed $\ba$, the classifier block of \eqref{eq:model} is exactly one elastic-net PinSVM with parameter $\tau_{\mathrm{eff}}$.
\end{proposition}

A proof is provided in the Supplementary Material. This proposition clarifies the role of the weights. They do not define separate classifiers. Instead, they induce a continuous parameter between the smallest and largest candidate parameters. Since $\tau_{\mathrm{eff}}\in[\min_g\tau_g,\max_g\tau_g]$, the outer weighting procedure can be interpreted as learning a parameter from a finite reference family while preserving a standard shared PinSVM classifier block.

\subsection{Candidate weighting and final selection}
\begin{figure*}[t]
\centering
\includegraphics[width=.92\textwidth]{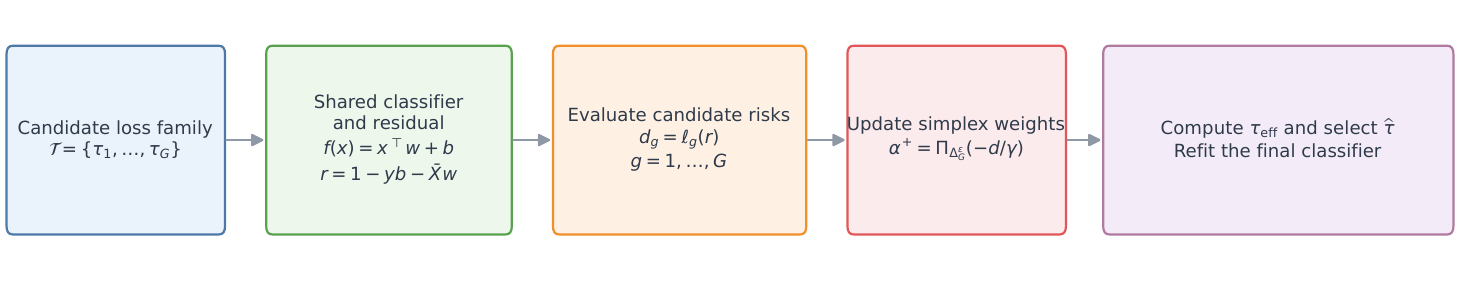}
\caption{Data-driven loss weighting with one shared classifier and one shared residual. The index $g$ labels only the candidate loss functions and their weights.}
\label{fig:workflow}
\end{figure*}

For fixed $(b,\bw,\br)$, define the candidate-risk vector
\[
\bm d(\br)
=\left(\ell_1(\br),\ldots,\ell_G(\br)\right)^\top.
\]
For $\gamma>0$, the weight subproblem is
\[
\min_{\ba\in\Delta_G^\varepsilon}
\bm d(\br)^\top\ba+\frac\gamma2\norm{\ba}_2^2,
\]
where $\bm d(\br)\in\R^G$ collects the $G$ candidate empirical risks. Let $\proj_{C}(\bm v)$ denote the Euclidean projection of a vector $\bm v$ onto a closed convex set $C$. Strong convexity gives the unique solution
\begin{equation}
\ba^+=\proj_{\Delta_G^\varepsilon}
\left(-\frac{\bm d(\br)}{\gamma}\right).
\label{eq:weight}
\end{equation}
The projection can be evaluated by a sorting-based simplex algorithm \cite{duchi2008projection}.
When $\gamma=0$, the same subproblem becomes a linear minimization over the simplex and may have nonunique solutions; this limiting case is used only for the oracle comparison below. For fixed $\ba$, Proposition~\ref{prop:effective} reduces the classifier update to one PinSVM at $\tau_{\mathrm{eff}}(\ba)$. The outer algorithm therefore alternates between one shared classifier fit and one simplex projection. Figure~\ref{fig:workflow} summarizes this logic.

\begin{algorithm}[t]
\caption{Data-driven pinball-loss selection}
\label{alg:outer}
\KwIn{$\bm X,\by,\cT,\lambda_1,\lambda_2,\gamma>0,\varepsilon$}
Set the outer-iteration index $s=0$; initialize $\alpha_g^0=1/G$ and a shared classifier $(b^0,\bw^0)$\;
\Repeat{the objective change and $\norm{\ba^{s+1}-\ba^s}$ are below tolerance}{
Compute $\tau_{\mathrm{eff}}^s$ from \eqref{eq:taueff}\;
Fit one shared elastic-net PinSVM at $\tau_{\mathrm{eff}}^s$ using Algorithm~\ref{alg:inner}\;
Set $\br^{s+1}=\one-\by b^{s+1}-\bar{\bm X}\bw^{s+1}$\;
Evaluate $d_g^{s+1}=\ell_g(\br^{s+1})$ for $g=1,\ldots,G$\;
Update $\ba^{s+1}$ using \eqref{eq:weight}\;
$s\leftarrow s+1$\;
}
Compute the final $\tau_{\mathrm{eff}}$ and select
$\widehat\tau\in\argmin_{\tau_g\in\cT}|\tau_g-\tau_{\mathrm{eff}}|$\;
Refit one standard elastic-net PinSVM at $\widehat\tau$ on the full training sample\;
\KwOut{$\ba$, $\tau_{\mathrm{eff}}$, $\widehat\tau$, and the refitted classifier}
\end{algorithm}

The refit step has a specific role. The converged shared classifier corresponds to the continuous $\tau_{\mathrm{eff}}$, whereas the reported discrete selection is $\widehat\tau\in\cT$. Refitting makes the final predictive model exactly the standard elastic-net PinSVM associated with the selected candidate. There is no candidate-dependent regularization scaling: the same $\lambda_1$ and $\lambda_2$ are used throughout, and $\alpha_g$ appears only in the weighted empirical loss.

\section{Vertically distributed feature-splitting solver}
\subsection{Shared-classifier subproblem}
At a given outer iteration, $\tau=\tau_{\mathrm{eff}}$ is fixed and the classifier block is
\begin{equation}
\begin{aligned}
\min_{b,\bw,\br}\quad
&\psi_\tau(\br)+\lambda_1\norm{\bw}_1
+\frac{\lambda_2}{2}\norm{\bw}_2^2,\\
\text{s.t.}\quad
&\by b+\bar{\bm X}\bw+\br=\one,
\end{aligned}
\label{eq:subproblem}
\end{equation}
where $\psi_\tau(\br)=n^{-1}\sum_iL_\tau(r_i)$. No candidate weight appears in \eqref{eq:subproblem}; its influence has already been summarized by $\tau_{\mathrm{eff}}$.

Let $\bm q\in\R^{p+1}$ collect the intercept and slope coefficients, let $\bm A\in\R^{n\times(p+1)}$ be the augmented label-weighted design matrix, and let $\phi$ denote the elastic-net regularizer:
\[
\bm q=(b,\bw^\top)^\top,
\qquad
\bm A=[\by,\bar{\bm X}],
\qquad
\phi(\bm q)=\lambda_1\norm{\bw}_1
+\frac{\lambda_2}{2}\norm{\bw}_2^2.
\]
Then \eqref{eq:subproblem} is
\[
\min_{\bm q,\br}\ \phi(\bm q)+\psi_\tau(\br)
\quad\text{s.t.}\quad
\bm A\bm q+\br=\one.
\]
Using the saddle function, where $\bu\in\R^n$ is the dual vector associated with the equality constraint,
\[
\mathcal L(\bm q,\br,\bu)
=\phi(\bm q)+\psi_\tau(\br)
-\bu^\top(\bm A\bm q+\br-\one),
\]
we apply a preconditioned primal-dual prediction--correction step. This step can be viewed within the maximal-monotone proximal-point framework \cite{rockafellar1976ppa,bauschke2017convex} and as a customized proximal-point realization for the associated saddle operator \cite{cai2013ppa,gu2014customized}. It is also closely related to standard primal--dual splitting \cite{chambolle2011firstorder,condat2013primaldual}. Its feature-block form follows the partition-insensitive structure in \cite{wu2025featureppa}.

\subsection{Distributed determination of the global proximal parameter}
Let $\mu>0$ denote the primal--dual scaling parameter and let $\eta>0$ denote the global proximal parameter shared by all workers. We write $\lambda_{\max}(\cdot)$ for the largest eigenvalue, $\norm{\cdot}_2$ for the spectral norm of a matrix or Euclidean norm of a vector, and $\norm{\cdot}_F$ for the Frobenius norm. The feature-block updates require
\begin{equation}
\eta>\mu\lambda_{\max}(\bm A^\top\bm A)
=\mu\norm{\bm A}_2^2,
\qquad
\bm A=[\by,\bar{\bm X}].
\label{eq:stepsize}
\end{equation}
A local value of $\eta$ at each worker would generally destroy the iterate-wise partition-insensitivity result. The parameter must therefore be determined from the complete design while keeping the raw feature blocks local.

Under the column partition $\bar{\bm X}=[\bar{\bm X}_1,\ldots,\bar{\bm X}_M]$, a certified one-round upper bound follows from
\begin{equation}
\norm{\bm A}_2^2
\leq \norm{\bm A}_F^2
=\norm{\by}_2^2+\sum_{m=1}^{M}\norm{\bar{\bm X}_m}_F^2
=n+\sum_{m=1}^{M}\norm{\bm X_m}_F^2.
\label{eq:etabound}
\end{equation}
Worker $m$ computes the scalar $s_m=\norm{\bm X_m}_F^2$ locally and sends only $s_m$ to the coordinator. For any safety factor $\delta>0$, the coordinator may set
\begin{equation}
\eta=(1+\delta)\mu
\left(n+\sum_{m=1}^{M}s_m\right),
\label{eq:etachoice}
\end{equation}
which satisfies \eqref{eq:stepsize}, is independent of the number and sizes of the feature blocks, and requires only one scalar from each worker. This choice is conservative but provides a fully verifiable global constant.

A tighter practical estimate can be obtained by distributed power iteration on
$\bm A\bm A^\top=\by\by^\top+\sum_m\bar{\bm X}_m\bar{\bm X}_m^\top$. Let $t$ denote the power-iteration index. Given a unit vector $\bm v^t\in\R^n$, worker $m$ returns
$\bm p_m^t=\bar{\bm X}_m(\bar{\bm X}_m^\top\bm v^t)$, after which the coordinator forms
\begin{equation}
\bm p^t=\by(\by^\top\bm v^t)+\sum_{m=1}^{M}\bm p_m^t,
\qquad
\bm v^{t+1}=\frac{\bm p^t}{\norm{\bm p^t}_2}.
\label{eq:distributedpower}
\end{equation}
This procedure never transmits raw columns and produces the same matrix--vector product as centralized power iteration. Because an unconverged Rayleigh quotient need not be an upper bound, the theoretical results use the certified choice \eqref{eq:etachoice}; the power estimate is used only to assess or reduce conservatism when accompanied by a valid safeguard. This distinction follows the global-constant principle in partition-insensitive parallel algorithms \cite{wu2025consensus,wu2025featureppa}.

\subsection{Column-partitioned updates}
Partition only the feature columns and slope coefficients:
\[
\bm X=[\bm X_1,\ldots,\bm X_M],
\qquad
\bw=(\bw_1^\top,\ldots,\bw_M^\top)^\top.
\]
The scalar $b$ remains at the coordinator. Choose $\mu>0$ and determine the common $\eta$ as described above. At inner iteration $k$, worker $m$ performs
\begin{equation}
\bw_m^{k+1}
=\frac{\soft_{\lambda_1}
\left(\eta\bw_m^k+(\bm Y\bm X_m)^\top\bu^k\right)}
{\eta+\lambda_2},
\label{eq:wupdate}
\end{equation}
where $\soft_c(v)=\operatorname{sign}(v)\max\{|v|-c,0\}$ componentwise. The coordinator updates the unpartitioned intercept by
\begin{equation}
b^{k+1}=b^k+\frac{\by^\top\bu^k}{\eta}.
\label{eq:bupdate}
\end{equation}
Each worker returns the partial margin
\[
\bm h_m^{k+1}=\bm Y\bm X_m\bw_m^{k+1}.
\]

For the residual step, let $z_i^k=r_i^k+u_i^k/\mu$ and
\[
c_\tau=\frac{1}{n\mu(1+\tau)}.
\]
The proximal mapping of the normalized pinball loss in \cite{liang2024ladmm} is
\begin{equation}
r_i^{k+1}
=\max\left\{z_i^k-c_\tau,
\min\left(0,z_i^k+\tau c_\tau\right)\right\}.
\label{eq:rupdate}
\end{equation}
Define the feasibility residual
\[
\bm e^{k+1}
=\by b^{k+1}+\sum_{m=1}^M\bm h_m^{k+1}
+\br^{k+1}-\one.
\]
The corrected dual update is
\begin{equation}
\bu^{k+1}
=\bu^k-\frac\mu2\left(2\bm e^{k+1}-\bm e^k\right).
\label{eq:uupdate}
\end{equation}
Equations \eqref{eq:wupdate}--\eqref{eq:uupdate} contain no candidate index. During one outer iteration, they solve one shared classifier at the current $\tau_{\mathrm{eff}}$.

\begin{algorithm}[t]
\caption{Vertically distributed solver for one shared PinSVM}
\label{alg:inner}
\KwIn{Coordinator: $\by,b^0,\br^0,\bu^0,\tau,\mu,\eta$; worker $m$: $\bm X_m,\bw_m^0$}
Set the inner-iteration index $k=0$; workers send initial partial margins, and the coordinator forms $\bm e^0$\;
\Repeat{the norm of successive primal--dual changes is below tolerance}{
Coordinator broadcasts $\bu^k$ and updates the scalar $b^{k+1}$ using \eqref{eq:bupdate}\;
\For{$m=1,\ldots,M$ \textbf{in parallel}}{
Worker $m$ updates $\bw_m^{k+1}$ using \eqref{eq:wupdate}\;
Worker $m$ returns $\bm h_m^{k+1}=\bm Y\bm X_m\bw_m^{k+1}$\;
}
Coordinator updates $\br^{k+1}$ using \eqref{eq:rupdate}\;
Coordinator aggregates $\sum_m\bm h_m^{k+1}$ and forms $\bm e^{k+1}$\;
Coordinator updates $\bu^{k+1}$ using \eqref{eq:uupdate}\;
$k\leftarrow k+1$\;
}
\KwOut{Shared $b^k$, residual $\br^k$, dual $\bu^k$, and coefficient blocks $\{\bw_m^k\}$}
\end{algorithm}

\subsection{Global loss evaluation under column partitioning}
\begin{figure*}[t]
\centering
\includegraphics[width=.96\textwidth]{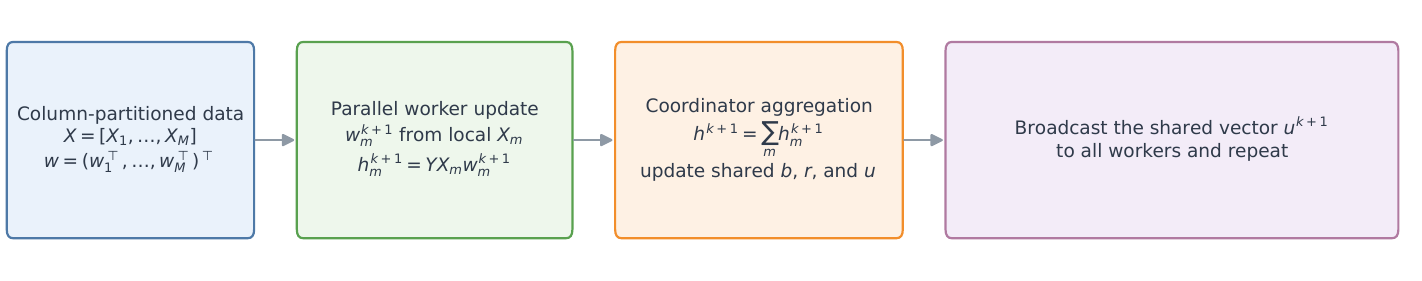}
\caption{Column-partitioned computation for the shared classifier. Workers update local slope blocks and return partial margins; the coordinator maintains the intercept, residual, and dual vector.}
\label{fig:architecture}
\end{figure*}

Column partitioning does not prevent the evaluation of the candidate losses on the full sample. At the end of an inner iteration, worker $m$ returns the sample-length partial margin
\[
\bm h_m=\bm Y\bm X_m\bw_m.
\]
The coordinator aggregates these vectors and reconstructs the global signed margin contribution
\[
\bm h=\sum_{m=1}^{M}\bm h_m
=\bar{\bm X}\bw.
\]
Because the scalar intercept $b$ is also maintained by the coordinator, the shared residual for the complete vertically partitioned data is available as
\begin{equation}
\br=\one-\by b-\sum_{m=1}^{M}\bm h_m.
\label{eq:globalresidual}
\end{equation}
The coordinator can therefore evaluate every candidate risk without accessing any raw feature block,
\begin{equation}
 d_g=\ell_g(\br)
 =\frac{1}{n}\sum_{i=1}^{n}L_{\tau_g}(r_i),
 \qquad g=1,\ldots,G,
\label{eq:globalrisks}
\end{equation}
apply the simplex projection in \eqref{eq:weight}, and compute $\tau_{\mathrm{eff}}$ from \eqref{eq:taueff}. No additional worker-to-worker communication is required, and the outer loss-weighting step uses the same residual that would be obtained from the centralized matrix $\bm X$. Figure~\ref{fig:architecture} summarizes the column-partitioned inner iteration; the global loss evaluation then uses the aggregated residual in \eqref{eq:globalresidual}.

\subsection{Communication and privacy scope}
Raw columns $\bm X_m$ and local coefficients $\bw_m$ remain at their sites. In each inner iteration, the coordinator broadcasts one vector $\bu^k\in\R^n$ to every worker and receives one partial margin $\bm h_m^{k+1}\in\R^n$ from every worker. The total uncompressed communication is therefore $2Mn$ floating-point values per iteration. The coordinator performs $O(nM)$ aggregation, while worker $m$ performs two dominant matrix--vector products with cost $O(np_m)$. The outer candidate-risk evaluation in \eqref{eq:globalrisks} is performed entirely at the coordinator and does not add communication.

Here $O(\cdot)$ denotes standard asymptotic arithmetic complexity. The architecture improves data locality because raw feature blocks are not transmitted. The proposed architecture does not provide a cryptographic privacy guarantee. Partial margins and dual vectors may still reveal information under adversarial models, and formal privacy would require secure aggregation, encryption, or differential-privacy mechanisms. The present contribution is therefore a vertically distributed optimization architecture, not a formal privacy protocol.

\section{Theoretical properties}
This section compares the proposed model with ordinary fixed-$\tau$ elastic-net PinSVMs and then summarizes the guarantees for the outer loss-weighting scheme, the inner variable-splitting algorithm, and the distributed column partition. Detailed proofs are provided in the Supplementary Material.

For each candidate $\tau_g$, define
\begin{equation}
\begin{aligned}
\mathcal Z&=\{(b,\bw,\br):\by b+\bar{\bm X}\bw+\br=\one\},\\
V_g&=\min_{(b,\bw,\br)\in\mathcal Z}
\left\{\ell_g(\br)+\lambda_1\norm{\bw}_1
+\frac{\lambda_2}{2}\norm{\bw}_2^2\right\}.
\end{aligned}
\label{eq:fixedvalue}
\end{equation}
The set $\mathcal Z$ collects the common margin-feasible triples used by both the proposed model and the fixed-candidate models. Let $(b_g^\star,\bw_g^\star,\br_g^\star)$ be a minimizer in the definition of $V_g$. Then $V_g$ is the optimal empirical objective of the ordinary elastic-net PinSVM with parameter fixed at $\tau_g$, and $\min_g V_g$ is the oracle benchmark over the candidate family.

\begin{theorem}[Oracle comparison with fixed-$\tau$ models]\label{thm:oracle}
Let $(\widehat b,\widehat{\bw},\widehat{\br},\widehat{\ba})$ be a global minimizer of \eqref{eq:model}, and define the classifier part of its objective by
\[
\widehat V=\sum_{g=1}^G\widehat\alpha_g\ell_g(\widehat{\br})
+\lambda_1\norm{\widehat{\bw}}_1
+\frac{\lambda_2}{2}\norm{\widehat{\bw}}_2^2.
\]
Choose $g^\star\in\argmin_g V_g$, set
\[
C_{g^\star}=\frac1n\sum_{i=1}^n|r_{g^\star,i}^\star|
\sum_{h\neq g^\star}|\tau_h-\tau_{g^\star}|,
\qquad
\rho_\varepsilon^2=\bigl[1-(G-1)\varepsilon\bigr]^2+(G-1)\varepsilon^2,
\]
where $r_{g^\star,i}^\star$ is the $i$th component of $\br_{g^\star}^\star$. Then
\begin{equation}
\widehat V
\leq \min_{1\leq g\leq G}V_g
+\varepsilon C_{g^\star}
+\frac{\gamma}{2}\left(\rho_\varepsilon^2-\norm{\widehat{\ba}}_2^2\right)
\leq \min_{1\leq g\leq G}V_g
+\varepsilon C_{g^\star}
+\frac{\gamma}{2}\left(\rho_\varepsilon^2-\frac1G\right).
\label{eq:oracle}
\end{equation}
In particular, when $\varepsilon=0$,
\[
\widehat V\leq \min_g V_g
+\frac\gamma2\left(1-\norm{\widehat{\ba}}_2^2\right)
\leq \min_gV_g+\frac\gamma2\left(1-\frac1G\right).
\]

\end{theorem}
When $\gamma=\varepsilon=0$, Theorem~\ref{thm:oracle} gives $\widehat V\leq\min_gV_g$, so a global minimizer is no worse in empirical classifier objective than the best fixed candidate. This is the sharpest oracle comparison, but it is not automatically the most stable computational choice. With $\gamma=0$, the weight update is linear over the simplex and typically selects a vertex; ties may produce nonunique solutions, and small perturbations can switch the selected vertex. A positive $\gamma$ makes the weight subproblem strongly convex, yielding a unique and smoother update, while a small $\varepsilon>0$ keeps every candidate weakly active and reduces boundary degeneracy. The explicit gap in \eqref{eq:oracle} therefore quantifies the trade-off introduced by these stabilizing choices. The case $\gamma=0$ is a model-level benchmark rather than the setting used by the projection update \eqref{eq:weight}; moreover, the alternating scheme guarantees stationarity, not attainment of a global minimizer.

\begin{theorem}[Descent and stationary accumulation points]\label{thm:outer}
Assume $\lambda_2>0$, $\gamma>0$, $\Delta_G^\varepsilon\neq\varnothing$, both classes occur in the training sample, the level set generated from the initialization is compact, and each classifier subproblem is solved exactly. Then Algorithm~\ref{alg:outer} produces a nonincreasing sequence of objective values for \eqref{eq:model}. Every accumulation point of the outer sequence $(b^s,\bw^s,\br^s,\ba^s)$ is a coordinatewise minimum and satisfies the first-order stationary conditions of the two-block constrained problem.
\end{theorem}
The theorem ensures that each exact outer update cannot increase the joint objective and that every accumulation point satisfies first-order stationarity. Since the joint problem is biconvex rather than jointly convex, the result does not claim global optimality.

\begin{theorem}[Convergence of the inner solver]\label{thm:inner}
Suppose the saddle-point set of \eqref{eq:subproblem} is nonempty and \eqref{eq:stepsize} holds. Then the sequence generated by Algorithm~\ref{alg:inner} converges to a saddle point $(\bm q^\star,\br^\star,\bu^\star)$. Moreover, let $\bm H\succ0$ be a positive-definite energy matrix specified in the Supplementary Material, and define $\norm{\bm v}_{\bm H}^2=\bm v^\top\bm H\bm v$. For every nonnegative integer $T$,
\begin{equation}
\min_{0\leq k\leq T}
\norm{\bm g^{k+1}-\bm g^k}_{\bm H}^2
\leq
\frac{\norm{\bm g^0-\bm g^\star}_{\bm H}^2}{T+1},
\label{eq:rate}
\end{equation}
where $\bm g^k=(\bm q^k,\br^k,\bu^k)$ collects all inner variables at iteration $k$.
\end{theorem}
Thus the iterates for the fixed-$\tau$ classifier subproblem converge to a saddle point. The bound in \eqref{eq:rate} is a best-iterate $O(1/T)$ estimate for the squared step residual and does not assert linear convergence.

\begin{theorem}[Partition insensitivity]\label{thm:partition}
Run Algorithm~\ref{alg:inner} with the same initialization, $\mu$, global $\eta$, and $\tau$ under any two column partitions of $\bm X$. In exact arithmetic, the coordinator sequences $(b^k,\br^k,\bu^k)$ are identical, and concatenating the worker blocks gives the same full coefficient vector $\bw^k$ at every iteration. In floating-point arithmetic, differences are limited to reduction-order rounding effects.
\end{theorem}
This result gives distributed equivalence: in exact arithmetic and under common initialization and global parameters, changing the number or sizes of the feature blocks affects the workload and communication pattern but not the mathematical iterates or their limit.

\section{Experiments}
\subsection{Setup}
We use Breast Cancer, Wine classes 0--1, Iris classes 1--2, and Digits classes 3--8 from scikit-learn \cite{pedregosa2011sklearn}. These tasks cover low-dimensional biomedical and botanical data, a chemically characterized dataset, and a moderate-dimensional image representation. Digits38-HD augments the 64 digit features with independent Gaussian noise variables to obtain $p=1000$; it is included to examine how elastic-net regularization and loss selection behave when informative coordinates are embedded in many irrelevant variables. The experiment is a controlled stress test rather than a claim that independent Gaussian variables reproduce every real high-dimensional noise mechanism.

Features are standardized using statistics computed from the training fold only. Every experiment uses a 70/30 stratified split and is repeated over five random seeds. Accuracy measures the final hard-label decision, whereas AUC evaluates the ranking induced by the signed decision score and is less dependent on a single classification threshold. The test fold is used only for final evaluation and is not involved in candidate weighting, parameter selection, or refitting. This separation is important because the learned loss weights and the selected pinball parameter are part of model training, not post hoc test-set choices.

The candidate set is
\[
\cT=\{0,0.1,0.25,0.5,0.75,1\}.
\]
It spans the hinge endpoint, several intermediate asymmetric losses, and the opposite endpoint with a deliberately coarse grid. Unless stated otherwise, $\lambda_1=\lambda_2=0.01$, $\gamma=1$, and $\varepsilon=10^{-4}$. These common values isolate the effect of loss selection; they are not presented as universally optimal hyperparameters. The quadratic weight regularizer prevents an unstable vertex solution when candidate risks are close, while the truncated-simplex floor keeps every candidate weakly active and reduces boundary degeneracy during alternating updates.

The same global $\eta$ is reused for every column partition. In the experiments, it is obtained from a converged power-method estimate of $\norm{\bm A}_2^2$ with a 1\% inflation factor. The convergence result applies whenever the resulting value satisfies \eqref{eq:stepsize}; Equation~\eqref{eq:etachoice} provides a certified alternative. We compare a linear SVM, an elastic-net hinge model ($\tau=0$), a fixed elastic-net PinSVM with $\tau=0.5$, and the proposed selected-$\tau$ refit. The first baseline represents a standard linear SVM implementation, the second isolates the effect of elastic-net regularization at the hinge endpoint, and the third tests whether data-driven selection improves upon one conventional interior pinball value. All experiments were conducted on a workstation equipped with an AMD Ryzen 9 7950X 16-core processor (4.50 GHz base clock) and 32 GB RAM\@. The numerical procedures were implemented in Python. The principal experimental settings are reported above, and a reference implementation of the shared-classifier solver and the column-partition equivalence test is provided with the supplementary files.

\subsection{Predictive comparison}
Table~\ref{tab:results} reports mean accuracy and AUC with standard deviations. The selected model is competitive but not uniformly dominant; accordingly, the results should be interpreted as evidence of adaptive loss selection rather than uniform superiority. It improves mean accuracy over the fixed $\tau=0.5$ model on Breast Cancer, Wine 0--1, and Iris 1--2, while the fixed model is slightly better on the two digit tasks. LinearSVM remains strongest on some lower-dimensional benchmark datasets. These results indicate that the weighting mechanism provides a data-dependent loss choice, not a guaranteed improvement over every baseline.

The largest accuracy change relative to the fixed pinball model occurs on Iris 1--2, where the selected model reaches $0.987$ compared with $0.967$. The corresponding AUC values are already close to one, so the practical difference is mainly reflected in the operating threshold rather than a dramatic change in ranking quality. On Wine 0--1, all methods attain essentially perfect AUC; the small accuracy differences should therefore be interpreted cautiously because the test folds are small and a single observation can noticeably affect the reported proportion.

The digit results illustrate the opposite case. Digits 3--8 appears nearly separable for a linear classifier, leaving little room for a loss-selection mechanism to improve the result. In Digits38-HD, elastic-net models outperform the unregularized linear baseline, suggesting that regularization is beneficial in this constructed noisy setting. However, the selected loss does not exceed the hinge or fixed-pinball alternatives in mean accuracy. This outcome is consistent with the diffuse weight profile reported below: the shared residual provides limited evidence that one normalized pinball geometry is decisively preferable. The method should therefore be viewed as an interpretable adaptive choice that can also report weak selection evidence, rather than as a procedure that guarantees higher accuracy.

\begin{table*}[t]
\centering
\caption{Mean $\pm$ standard deviation over five stratified splits.}
\label{tab:results}
\small
\begin{tabular}{llcc}
\toprule
Dataset & Method & Accuracy & AUC \\
\midrule
\multirow{4}{*}{Breast Cancer}
& LinearSVM & $0.973\pm0.009$ & $0.993\pm0.007$ \\
& EN-HingeSVM & $0.951\pm0.016$ & $0.990\pm0.005$ \\
& EN-PinSVM ($\tau=0.5$) & $0.946\pm0.011$ & $0.989\pm0.005$ \\
& Selected-$\tau$ & $0.954\pm0.011$ & $0.991\pm0.005$ \\
\midrule
\multirow{4}{*}{Wine 0--1}
& LinearSVM & $0.979\pm0.021$ & $1.000\pm0.000$ \\
& EN-HingeSVM & $0.985\pm0.014$ & $1.000\pm0.000$ \\
& EN-PinSVM ($\tau=0.5$) & $0.985\pm0.023$ & $1.000\pm0.000$ \\
& Selected-$\tau$ & $0.990\pm0.023$ & $1.000\pm0.000$ \\
\midrule
\multirow{4}{*}{Iris 1--2}
& LinearSVM & $0.960\pm0.037$ & $0.989\pm0.014$ \\
& EN-HingeSVM & $0.960\pm0.037$ & $0.995\pm0.005$ \\
& EN-PinSVM ($\tau=0.5$) & $0.967\pm0.033$ & $0.997\pm0.004$ \\
& Selected-$\tau$ & $0.987\pm0.018$ & $0.998\pm0.004$ \\
\midrule
\multirow{4}{*}{Digits 3--8}
& LinearSVM & $0.989\pm0.010$ & $1.000\pm0.000$ \\
& EN-HingeSVM & $0.974\pm0.012$ & $0.999\pm0.001$ \\
& EN-PinSVM ($\tau=0.5$) & $0.974\pm0.012$ & $0.999\pm0.002$ \\
& Selected-$\tau$ & $0.970\pm0.010$ & $0.998\pm0.002$ \\
\midrule
\multirow{4}{*}{Digits38-HD}
& LinearSVM & $0.944\pm0.022$ & $0.985\pm0.015$ \\
& EN-HingeSVM & $0.967\pm0.012$ & $0.994\pm0.007$ \\
& EN-PinSVM ($\tau=0.5$) & $0.959\pm0.011$ & $0.990\pm0.012$ \\
& Selected-$\tau$ & $0.957\pm0.011$ & $0.991\pm0.011$ \\
\bottomrule
\end{tabular}
\end{table*}

\begin{figure*}[t]
\centering
\begin{minipage}[t]{.49\textwidth}
\centering
\includegraphics[width=\linewidth]{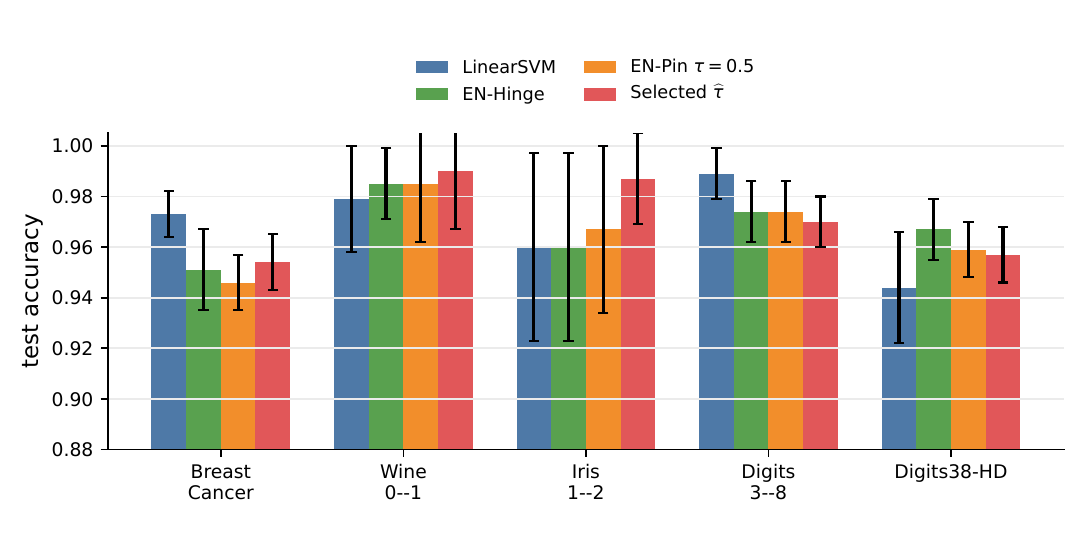}
\end{minipage}\hfill
\begin{minipage}[t]{.49\textwidth}
\centering
\includegraphics[width=\linewidth]{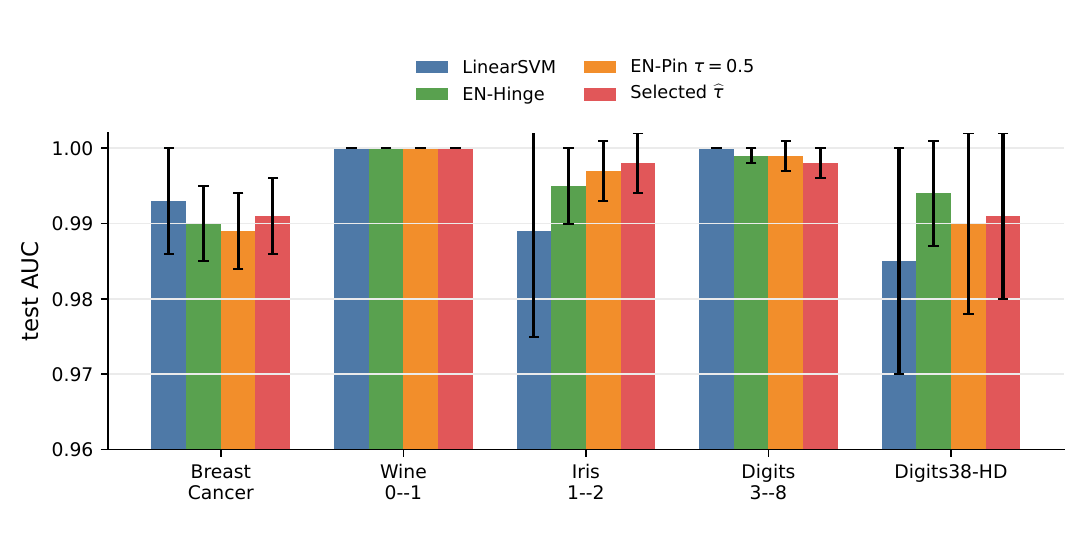}
\end{minipage}
\caption{Predictive comparison: test accuracy (left) and test AUC (right). Error bars show one standard deviation.}
\label{fig:predictive}
\end{figure*}

\subsection{Learned weights and selected parameters}
\begin{figure*}[t]
\centering
\begin{minipage}[t]{.49\textwidth}
\centering
\includegraphics[width=\linewidth]{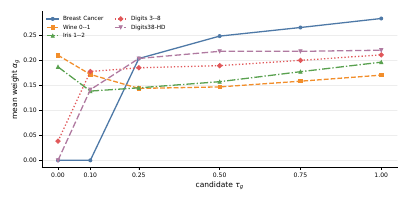}
\end{minipage}\hfill
\begin{minipage}[t]{.49\textwidth}
\centering
\includegraphics[width=\linewidth]{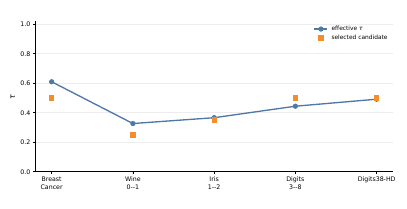}
\end{minipage}
\caption{Mean candidate weights (left) and effective/selected parameters (right).}
\label{fig:tauanalysis}
\end{figure*}

Figure~\ref{fig:tauanalysis} reports the mean weights and the resulting parameters. The weights are regularized by $\gamma$ and remain distributed rather than collapsing to a single vertex. Breast Cancer assigns gradually increasing mass to larger $\tau_g$, whereas Wine 0--1 favors smaller values. The other datasets produce nearly balanced profiles. These patterns yield effective parameters between approximately $0.335$ and $0.378$.

The selected candidate is $0.25$ for all five runs on Wine, Iris, Digits 3--8, and Digits38-HD\@. Breast Cancer alternates between $0.25$ and $0.5$ because its effective value lies near their midpoint. The nearly uniform Digits38-HD weights are informative: the shared residual does not strongly distinguish among the normalized candidate losses, so the method reports weak selection evidence rather than an unwarranted concentrated choice.

The effective parameter is not the arithmetic mean of the candidate values. It is determined by the positive and negative slopes of the normalized losses, so candidates with the same simplex weight can contribute differently after normalization. This explains why a broadly distributed weight vector may still yield a stable effective value. The subsequent nearest-grid projection serves only to report and refit a standard candidate model; the continuous effective parameter remains the more faithful summary of the learned loss geometry. In applications where a continuous parameter is acceptable, the refit could instead use the effective value directly.

\subsection{Outer convergence and partition insensitivity}
\begin{figure*}[t]
\centering
\begin{minipage}[t]{.49\textwidth}
\centering
\includegraphics[width=\linewidth]{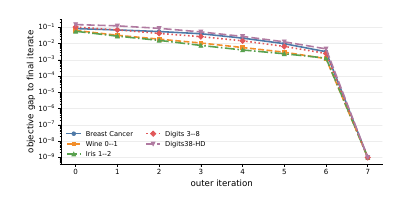}
\end{minipage}\hfill
\begin{minipage}[t]{.49\textwidth}
\centering
\includegraphics[width=\linewidth]{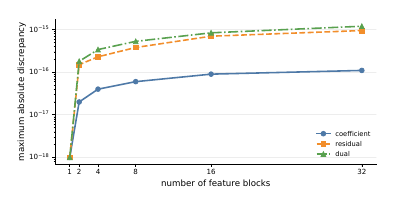}
\end{minipage}
\caption{Outer objective gaps (left) and discrepancies across feature partitions (right).}
\label{fig:optimization}
\end{figure*}

The left panel of Figure~\ref{fig:optimization} shows the objective gap relative to the final outer iterate. The curves decay rapidly and are consistent with the monotonicity result in Theorem~\ref{thm:outer}. Because the classifier block is solved numerically, tiny nonmonotone effects can occur if the inner tolerance is loose; the reported runs use a sufficiently tight tolerance to make the expected descent visible.

For partition verification, Digits38-HD is solved for 200 inner iterations with identical initialization and global $\eta$ under $1,2,4,8,16,$ and $32$ column blocks. The right panel reports the maximum discrepancy from the one-block sequence. Coefficient differences remain below $1.2\times10^{-16}$, while residual and dual discrepancies remain around $10^{-15}$, matching floating-point reduction error and supporting Theorem~\ref{thm:partition}.

This experiment tests iterate equivalence rather than merely agreement of final objective values. The same initialization, stopping horizon, and global proximal parameter are used for every partition, so a visible discrepancy would indicate that the mathematical update changed with the block layout. The observed machine-precision differences instead arise from the order in which floating-point partial margins are summed. Partition insensitivity is therefore an algorithmic reproducibility property; it does not imply that runtime, memory traffic, or communication latency are themselves independent of the partition.

\subsection{Parallel-environment evaluation}
We additionally evaluate the column-splitting solver in a multi-process environment. A synthetic high-dimensional problem with $n=1000$ and $p=20000$ is divided among $1,2,$ and $4$ worker processes. Each process is restricted to one BLAS thread, and each configuration performs 30 inner iterations over four repetitions. The effective parameter is fixed at $\tau=0.35$, close to the empirical values learned on the real datasets. The benchmark measures the solver realization only and does not simulate a wide-area institutional network.

\begin{figure*}[t]
\centering
\includegraphics[width=.98\textwidth]{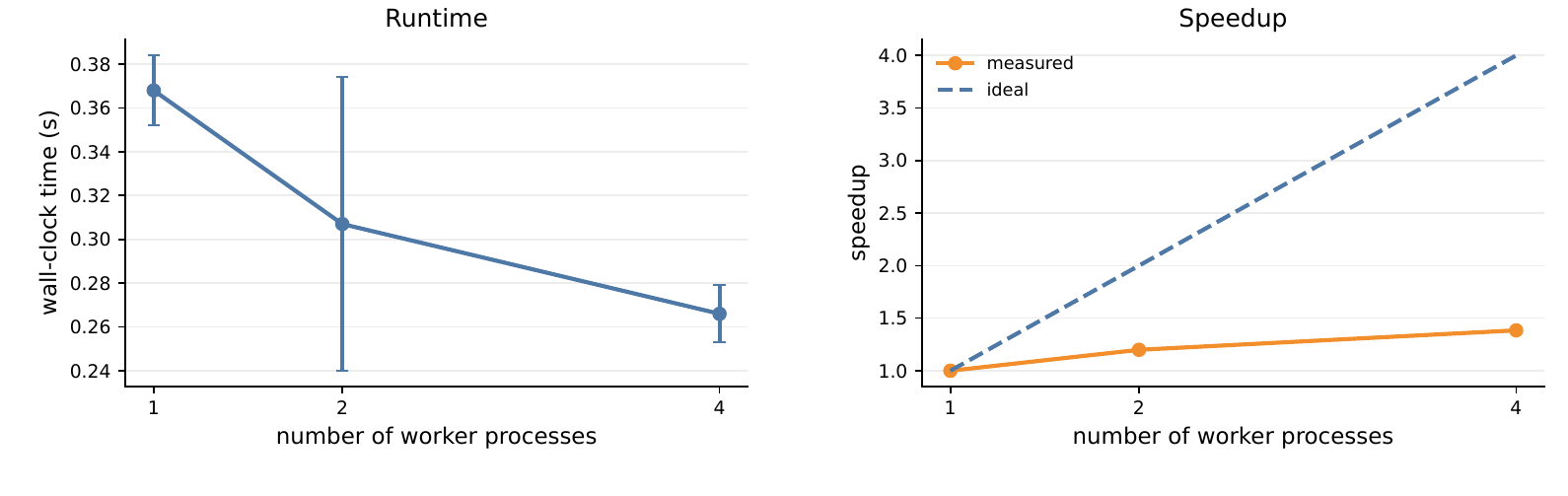}
\caption{Parallel runtime and speedup under one-, two-, and four-process column partitions. The left panel reports wall-clock time with one-standard-deviation error bars over four repetitions, and the right panel compares the measured speedup with the ideal linear baseline.}
\label{fig:parallel7}
\end{figure*}

\begin{figure*}[t]
\centering
\includegraphics[width=.98\textwidth]{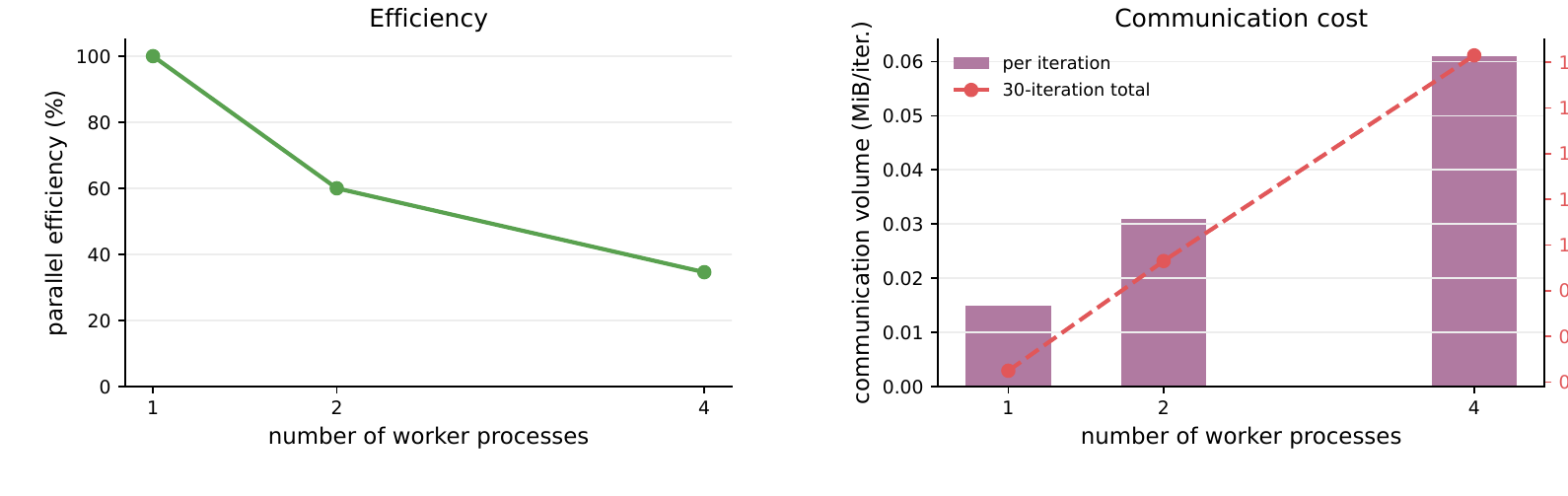}
\caption{Communication-side behavior of the multi-process implementation. The left panel reports parallel efficiency, and the right panel reports both the communication volume exchanged per iteration and the total exchanged volume over the 30-iteration run.}
\label{fig:parallel8}
\end{figure*}

Table~\ref{tab:parallel}, Figure~\ref{fig:parallel7}, and Figure~\ref{fig:parallel8} show the results. Runtime decreases from $0.368$ seconds with one process to $0.266$ seconds with four processes, corresponding to a measured speedup of $1.384$. Figure~\ref{fig:parallel7} separates the wall-clock runtime trajectory from the speedup curve so that the measured gain can be compared directly with the ideal linear baseline. The gain is sublinear because process communication, process scheduling, memory-bandwidth contention, and coordinator aggregation become increasingly important.

Figure~\ref{fig:parallel8} gives a more detailed communication-side view. The left panel reports the decline in parallel efficiency from $100.0\%$ to $34.6\%$. The right panel reports the communication volume per iteration and the total exchanged volume over the 30-iteration run, which grows from $0.45$ MiB with one worker to $1.83$ MiB with four workers. This behavior does not contradict the feature-splitting theory. The theory states that the numerical iterates do not depend on the column partition; it does not assert ideal hardware scaling. With only $n=1000$ sample-length messages, fixed process-management overhead is a non-negligible part of these short runs. Larger feature blocks or persistent distributed workers would increase the computation-to-communication ratio, whereas a wide-area deployment would introduce network latency absent from this workstation benchmark. The reported experiment should thus be read as evidence that the block updates can execute concurrently, not as a comprehensive systems evaluation.

\begin{table}[t]
\centering
\caption{Multi-process evaluation on the synthetic high-dimensional task.}
\label{tab:parallel}
\small
\begin{tabular}{ccccc}
\toprule
Workers & Time (s) & Speedup & Efficiency & MiB/iter. \\
\midrule
1 & $0.368\pm0.016$ & 1.000 & 1.000 & 0.015 \\
2 & $0.307\pm0.067$ & 1.199 & 0.600 & 0.031 \\
4 & $0.266\pm0.013$ & 1.384 & 0.346 & 0.061 \\
\bottomrule
\end{tabular}
\end{table}

\section{Discussion and conclusion}
The proposed model is a data-driven loss-parameter selector built around one shared elastic-net SVM\@. The intercept, slope vector, and residual are common to all candidates; the index $g$ labels the loss family $\ell_g$, whose members use parameters $\tau_g$ and weights $\alpha_g$. At a global minimizer, Theorem~\ref{thm:oracle} bounds the proposed model's empirical classifier objective by that of the best fixed candidate plus explicit weight-regularization and simplex-truncation terms. Because every normalized loss is evaluated on the same residual, their weighted sum admits an exact effective-$\tau$ representation. This makes the learned weights interpretable while avoiding unnecessary classifier duplication, residual variables, and repeated regularization terms.

The distributed contribution has a complementary scope: only feature columns and corresponding slope blocks are partitioned. The intercept remains a single coordinator scalar, and workers exchange only partial margins and one shared dual vector. Under a common global proximal constant, concatenated distributed iterates coincide with centralized iterates in exact arithmetic. The multi-process experiment shows a measured multi-process speedup, although communication and coordination prevent linear scaling and the architecture alone does not provide a formal privacy guarantee.

The empirical findings warrant a cautious interpretation. The selected loss improves upon the fixed $\tau=0.5$ model on several datasets but is not uniformly superior, and nearly uniform weights on Digits38-HD reveal weak evidence for distinguishing normalized candidates. Larger image, text, and biomedical benchmarks, nested tuning of $\lambda_1$, $\lambda_2$, and $\gamma$, and deployment over genuinely separated institutions remain important future work. Within these limits, the paper provides a data-driven parameter-selection model, a column-partitioned implementation, and theoretical guarantees consistent with the shared-classifier formulation.

\section*{Supplementary material}
Detailed proofs, the simplex-projection procedure, the computational and communication complexity analysis, and the multiclass extension are provided in the accompanying Supplementary Material.

\section*{Funding}
This work was supported by the National Natural Science Foundation of China [grant number 12401664] and the Natural Science Foundation of Chongqing [grant number CSTB2024NSCQ-MSX0855]. Xiaofei Wu also acknowledges support from the Visiting Scholar Program of the Chern Institute of Mathematics. The funding bodies had no role in the study design, data analysis, interpretation of the results, preparation of the manuscript, or decision to submit the work for publication.

\section*{CRediT authorship contribution statement}
\textbf{Xiaofei Wu:} Conceptualization, Methodology, Supervision, Project administration, Writing--review and editing. 
\textbf{Kai Qi:} Conceptualization, Methodology, Formal analysis, Validation, Writing--original draft, Funding acquisition. 
\textbf{Rongmei Liang:} Conceptualization, Methodology, Software, Data curation, Visualization, Writing--original draft, Writing--review and editing. 
Xiaofei Wu and Kai Qi contributed equally to this work. All authors approved the final manuscript.

\section*{Data and code availability}
The benchmark datasets are publicly available through scikit-learn \cite{pedregosa2011sklearn}. No restricted or proprietary data were used.A reference implementation of the shared-classifier solver, the Digits38-HD construction, and the column-partition equivalence test is publicly available at \url{https://github.com/xfwu1016/DP-ENSVM}.

\section*{Declaration of competing interest}
The authors declare that they have no known competing financial interests or personal relationships that could have appeared to influence the work reported in this paper.

\section*{Declaration of generative AI and AI-assisted technologies in the manuscript preparation process}
During the preparation of this work, the authors used ChatGPT (OpenAI) to improve English-language readability and LaTeX formatting. After using this tool, the authors reviewed and edited the content as needed and take full responsibility for the content of the publication.

\bibliographystyle{elsarticle-num}
\bibliography{references}
\end{document}